\documentclass{article}
\pdfoutput=1
\usepackage{amsmath}
\usepackage{times}
\usepackage{graphicx}
\usepackage{subcaption}
\usepackage{natbib}
\usepackage{algorithm}
\usepackage{algorithmic}
\usepackage{hyperref}

\renewcommand{\cite}[1]{\citep{#1}}
\setcitestyle{authoryear,round,citesep={;},aysep={,},yysep={;}}

\newif\ificml
\icmlfalse

\ificml

\usepackage{icml2016} 

\else
\twocolumn
\fi

\newcommand{\mc}[1]{\mathcal{#1}}
\newcommand{\prob}{\text{Prob}}

\def\argmax{\operatorname{argmax}}

\usepackage{yub}

\begin{document}

\ificml

\twocolumn[

\icmltitle{TAPAS: Two-pass Approximate Adaptive Sampling for Softmax}


\icmlauthor{Yu Bai, Sally Goldman, Li Zhang}{}
\icmladdress{Google}

\icmlkeywords{adaptive sampling, softmax}

\vskip 0.3in
]

\else

\title{TAPAS: Two-pass Approximate Adaptive Sampling for Softmax}

\author{
\begin{tabular}{ccc}
    Yu Bai\footnote{The work was done during an internship at Google.} & Sally Goldman & Li Zhang\\
	\small{yub@stanford.edu} & \small{sgoldman@google.com} & \small{liqzhang@google.com}
\end{tabular}}

\date{Google Inc.}

\maketitle
\fi

\begin{abstract}
TAPAS is a novel adaptive sampling method for the softmax model. It
uses a two pass sampling strategy where the examples used to
approximate the gradient of the partition function are first sampled
according to a squashed population distribution and then resampled
adaptively using the context and current model. We describe an
efficient distributed implementation of TAPAS. We show, on both
synthetic data and a large real dataset, that TAPAS has low
computational overhead and works well for minimizing the rank loss for
multi-class classification problems with a very large label space.
\end{abstract}

\section{Introduction}\label{sec:intro}

Multi-class classification problems are ubiquitous in machine
learning: given empirical observations of pairs of context features
$x_i\in\mc{X}$ and discrete label $y_i\in[V]=\{1,\dots, V\}$, we wish
to learn to predict the label $y$ for any given $x$. Many tasks in
computer vision, natural language processing, and recommender systems
are by nature multi-class problems.

A particularly effective method for the multi-class classification
task is to model the conditional probability of $\prob[y|x]$ through a
neural network softmax model. In such a model, $\prob[y|x]$ is set to
be proportional to $\exp(\phi(x) \cdot \psi(y))$, where $\phi, \psi$
are parameterized functions that map each context and label to some
high dimensional space, called the context embedding and the label embedding, respectively. The model parameters of $\phi, \psi$ are then
learned by minimizing the empirical cross entropy loss using the
gradient descent method.

One challenge underlying this approach is that the vocabulary size $V$
can be very large as we apply the method to increasingly larger
tasks. For example, ImageNet \cite{RussakovskyDSKS15} consists of
around $10^4-10^5$ tags for images. In a language model, the
vocabulary of all words and common phrases can have $10^5-10^6$
entries. In a video recommendation task, $V$ is the number of videos
and is often on the order of $10^7-10^9$. Gradient-based training of
the softmax model requires computing the partition function
$Z(x)=\sum_{z\in [V]}\exp(\phi(x)\cdot\psi(z))$ at every training
step. When $V$ is large, computing $Z(x)$ becomes prohibitively
expensive.

Sampling based methods, such as importance sampling (also called
sampled softmax)~\cite{BengioS08} and noise contrastive
estimation~\cite{GutmannH12}, are common techniques to address this
problem. In such methods, at each training step, a small subset of
samples of $[V]$ is used to approximate the gradient of $Z(x)$. The
effectiveness of the sampling based method crucially depends on the
sampling distribution and the sample size. It also has to be done
efficiently to avoid large computational overhead.  The common
approach is to sample according to a pre-determined distribution,
usually dependent on the empirical distribution of the
labels~\cite{ChenGA16,JozefowiczVSSW16}.

In this paper, we propose
a \textbf{T}wo-pass \textbf{Ap}proximate \textbf{A}daptive \textbf{S}ampling
method (TAPAS) for the efficient training of the softmax model with
large vocabulary size. In TAPAS, the sampling is done in two
passes. In the first non-adaptive pass, we sample a subset $S'\subset
[V]$ according a pre-determined distribution, similar to the sampled softmax. In the second adaptive pass, we
resample a smaller set $S$ from $S'$ which are ``close'' to the
contexts, i.e. with higher predicted probability for the given
contexts, in the training batch. We then use $S$ for computing the
gradient updates on the model parameters.

Compared to the existing approaches, TAPAS chooses the samples
according to both the context and the current model parameters. The
resampling reduces the size of the samples so it is more efficient to
compute the gradients. 

Another useful view on the sampling is that the samples serve as
``negative'' labels since the gradient descent would cause the context
embedding and the label embeddings of the sampled classes to move away
from each other. Indeed, such sampling is also called negative
sampling. By focusing on a subset of the samples with higher logits,
the training procedure of TAPAS pays more attention to the ``hard''
negative labels, i.e. the classes likely to be confused with the true
label given the context. This leads to more efficient training and
better ranking accuracy such as the average precision score, similar
to~\cite{WestonBU11}.

The adaptive sampling, however, does incur computational overhead.  To
reduce the overhead, we present a distributed approximate sampling
algorithm that utilizes both the GPUs and the parallelism supported
by the state-of-the-art distributed machine learning platforms.  We
implemented TAPAS on Tensorflow~\cite{AbadiBCCDDDGIIK16} and show it
has very small overhead. We demonstrate the success of TAPAS on
both synthetic data and on a large scale real data set.

We provide empirical study of TAPAS in this paper. We conjecture that,
similar to the analysis showing that the Warp sampling of Wsabie
optimizes precision at $k$ versus optimizing the mean
rank~\cite{WestonBU11}, the adaptive sampling of TAPAS is closer to
optimizing a rank loss versus the full softmax loss.  However,
theoretical analysis proving this conjecture has been surprisingly
challenging and is a good direction for future work.

\subsection{Related work}\label{sec:related}

In this section we review related work. We list a variety of
techniques that have been proposed to address the prohibitive cost of
computing the negative gradient (or equivalently computing the
partition function and its gradient) when the vocabulary is extremely
large. For a good survey, see~\cite{ChenGA16}.

An important distinction which is
relevant to our work is the extent to which the methods depend on the {\em vocabulary} (e.g. the label frequencies),
the {\em context} (e.g., words before the word to be predicted in a language model) and the {\em model}
(e.g. current weights) itself.

\paragraph{Sampled softmax.} TAPAS builds on the idea of
sampling-based approximations of the softmax loss and its
gradients. These sampled softmax strategies specify a sampling distribution $Q$ from which they draw a subset of the label space $[V]$. Popular sampling distributions include the naive uniform distribution, frequency-based unigram (sample frequency) or
bigram distributions~\cite{BengioS03,BengioS08}, or
a power-raised distribution of the unigram~\cite{MilkolovCCD13, JiVSAD15}.
 These distributions are specified beforehand and do not adapt to the training process. TAPAS can be applied on top of any of such sampling schemes to add an adaptive layer to provide harder negatives.

More similar to TAPAS are the many variants of
{\em Adaptive Sampling} that adapt the sampling distribution $Q$ to the model training process. One work closely related to ours is the pioneering  method of {\em Adaptive Importance Sampling}~\cite{BengioS08}. \citeauthor{BengioS08} observe that sampling from the exponentiated logits $P(y|x)\propto\exp(\phi(x)\cdot\psi(y))$ will give us unbiased estimates of the full softmax gradient. To overcome the inefficiency of such a distribution, they define an approximate $Q$ using an
$n$-gram model that is a mixture of
a set of $k$-gram models which can be efficiently adapted during
training so that fewer examples are needed to approximate the gradient. 
In addition they introduce effective sample size (ESS) which
adaptively selects the size of the sample to use for the negative
sampling. An important limitation of their work is that their $Q$ has
an $n$-gram structure that is most appropriate for language models. In
contrast, TAPAS utilizes context and model information without
imposing structural assumptions so is suitable for more
tasks. Also ESS, while using the variance of the prediction probability to determine the sample size, does not subsample
it to use harder negatives and thus does not lend itself as well for ranking tasks.

\cite{JeanCMB14} introduce sampled softmax for neural machine
translation with very large vocabularies where the negative sampling
is performed in mini-batches.  However, the sampling method
does not depend on the current model, which is a key aspect of our work.

\paragraph{Tree-based methods.} Hierarchical Softmax (HSM)~\cite{MorinB05,Goodman01} is another popular
technique that organizes the
labels into a tree where the leaves are
the labels and the intermediate nodes are latent
variables. The probability
of a label is the product of the probabilities
of the latent variables along the path from the root
to the leaf. This decomposition allows a sequential computation of the
probabilities and saves the cost of computing the full
partition function $Z(x)$. The most common use of this is a two-level
HSM such as in~\cite{MilkolovCCD13}. HSM is most suitable when the
labels naturally forms a concept tree such as language models and is
able to achieve state-of-the-art perplexities on such tasks
\cite{JozefowiczVSSW16}. However, inferring a tree structure for a
general-purpose multi-class task might be highly non-trivial, and it is also hard to adjust the tree structure during training.

\paragraph{Efficient implementations.} Similar to our work which provides a very
efficient implantation on a distributed architecture such as
Tensorflow~\cite{AbadiBCCDDDGIIK16},
\cite{GraveJCGJ16} introduced an efficient softmax approximation that
is appropriate for distribution on GPUs. They
define a strategy to produce an approximate hierarchical model that is
well suited to efficient computation by GPUs.  Again, here the sampling
does not depend on the current model.

\paragraph{Other loss functions.} Many other loss functions prove successful in multi-class problems. One alternative approach is Noise Contrastive Estimation (NCE)~\cite{GutmannH12}.
These methods do not compute the negative gradient but instead
learn to discriminate between true labels and samples from a noise distribution. It essentially relates a multi-class problem to a binary problem. This is very suitable in a multi-label scenario, i.e. each context having multiple true labels. Another
approach are {\em Infrequent Normalization (Self Normalization)} that
perform infrequent updates of the negative gradient~\cite{AndreasK15}.

Although a very different direction, there is an interesting relationship between
Wsabie~\cite{WestonBU11} that uses stochastic gradient descent to optimize a ranking loss. The interesting
aspect about Wsabie is that the selection of the negative samples is very tightly linked
to the current model and it has been shown to improve the loss of the top ranked items
as compared to optimizing the AUC.  We show that the second phase of TAPAS achieves
a similar goal.

\subsection{Outline}\label{sec:outline}
The rest of this paper is organized as follows. In
Section~\ref{sec:prelim} we review preliminaries on softmax
regression, sampled softmax, rank losses, and motivate adaptive
sampling strategies. Section~\ref{sec:algo} describes the TAPAS
algorithm in detail. We further discuss issues on its computational
cost and provide an efficient implementation in
Section~\ref{sec:impl}. We demonstrate the success of TAPAS on
synthetic datasets and a large-scale real dataset in
Section~\ref{sec:exper}.

\section{Preliminaries}\label{sec:prelim}

 \paragraph{Notation.} We use $[V]$ to denote the set $\{1,2,\cdots,
 V\}$. Given a finite set $S$ and a function $f:S\to \R$, we use
 $\argmax^n_{x\in S} f(x)$ to denote the $n$ element subset of $S$
 that has the largest $f$ value. For two vectors $u, v\in \R^d$, let
 $u\cdot v$ denote their dot product.

 \paragraph{Multi-class classification.} We consider the multi-class
classification problem of predicting the label\footnote{All the
discussion in the paper directly generalizes to the case when each
context may receive multiple labels or a distribution of labels by using the cross-entropy loss with respect to the distribution of label classes. For
the simplicity of presentation, we focus on the case when there is a
single label for each context.} given a context where the label comes
from vocabulary set $[V]$. One classical example is the language
model where we predict a word from the context surrounding the word in
a sentence. It can also be used to model a recommendation system where
the context represents the user features, such as demographic
information and the past user activities, and each label represents an
item, for example a song or a video, that the user might like.

 \paragraph{Softmax regression.}  In the softmax regression, each context $x$ is
 mapped to a real feature vector $\phi(x)\in \R^d$, and each label $y$ is mapped to
 $\psi(y)\in \R^d$. Here $\phi$ is a neural network with
 multiple (non-linear) layers, and $\psi$ maps each id to a vector in
 $\R^d$. We model the conditional probability $\prob[y|x]$ as
 $\prob[y|x] \propto \exp(\phi(x) \cdot \psi(y))$, i.e.
\[\prob[y|x] = \exp(\phi(x) \cdot \psi(y))/Z_{\phi,\psi}(x)\,,\]

where $Z_{\phi,\psi}(x)=\sum_{z\in [V]} \exp(\phi(x)\cdot \psi(z))$ is
the {\em partition function\/} at $x$. We omit $\phi,\psi$ from the
notation when it is clear from the context. Given the data set $D$
containing empirical observation of $(x,y)$ pairs, the cross-entropy
loss (or softmax loss) of a model $(\phi,\psi)$ is defined as
\[L(\phi,\psi) = \sum_{(x,y)\in D} -\log\prob[y|x]\,.\]

The model parameters are then learned by minimizing the the above loss
$L$ over $(\phi,\psi)$ using the stochastic gradient descent
method. The point-wise gradient of the loss on an example $(x,y)\in D$ is:
\begin{align*}
&{}\nabla_{\theta}(-\log\prob[y|x])\\
= &{}\nabla_{\theta}[-(\phi(x) \cdot \psi(y)) + \log\sum_{z\in [V]} \exp(\phi(x) \cdot \psi(z))]\\
= &{}-\nabla_{\theta}(\phi(x) \cdot \psi(y)) + \sum_{z\in [V]} \prob[z|x] \nabla_{\theta}(\phi(x) \cdot \psi(z))\,.
\end{align*}

Applying to the context and the label embeddings, respectively, we have

\begin{align}
\nabla_{\phi(x)} & = -\psi(y) + \sum_{z\in[V]}\prob[z|x] \psi(z)\,,\label{eq:grad}\\
\nabla_{\psi(z)} & = (-\delta_{yz} + \prob[z|x])\phi(x)\,.
\end{align}

Here $\delta_{yz}$ denotes the Kronecker delta. By stochastic gradient
descent, we compute the mean of the point-wise gradient on a random
batch of examples and then apply the gradient descent.

With the trained model, the inference is done by computing
\begin{equation*}
\what{y}(x) = \argmax_{y\in[V]} \{\phi(x)\cdot \psi(y)\}\,.
\end{equation*}
It is common to compute the top $k$ labels if multiple candidates
are allowed.

\paragraph{Sampled softmax.}

By (\ref{eq:grad}), computing the gradient on $\phi(x)$ requires to
compute $\sum_{z\in[V]}\prob[z|x]\psi(z)$. This computation can be
prohibitively expensive for a large $V$. One solution is to sample a
subset $S\subset [V]$ to approximate $\nabla_{\phi(x)}$. In sampled softmax, each label $z$ in
 $[V]$ is assigned a probability $q_z$ for being selected in the
 sample. At each training step, a random subset $S$ of $[V]$ is
 sampled according to $q$, and the subset is used to approximate
 $\nabla L$. For computing $q_z$, one popular method is to use
 squashed empirical frequency. Suppose $f_z$ is the empirical
 frequency of the class $z$. We set
 $p_z \propto \max(f_z^\alpha, \beta)$ where $0\leq \alpha\leq 1$ is a squash
 exponent, and $\beta>0$ is a lower bound to guarantee a non-vanishing
 sampling. Sampled softmax and its variants have shown to produce the
 best accuracy for many multi-class classification
 problems~\cite{JozefowiczVSSW16,ChenGA16}.  Typically the sampled
 softmax uses the same sampling distribution for all the
 contexts. In~\cite{BengioS08}, it is suggested to use adaptive
 sampling  according to a separate model which makes rough estimation of the label probability.

From (\ref{eq:grad}), we can also see that the gradient descent would
move $\phi(x)$ towards $\psi(y)$ but away from $\psi(z)$ for $z\neq
y$. This is also why such sampling is commonly called negative
sampling as the sampled $z$ has the effect similar to a negative class
label.

\paragraph{Rank loss.} While the softmax loss is smooth and suitable for 
 minimization using gradient descent, in practice, the rank loss is
 often used for evaluating the model quality since usually only top
 model predictions are relevant to the applications. There are various
 variants of rank losses. In this paper we consider the common metrics
 of the precision and the mean average precision (MAP)
 metrics~\cite{KaggleWiki}, defined as follows. Suppose that the model
 produces a ranked list of predictions $z_1, z_2, \cdots$, and the
 true labels is a set $Y=\{y_1, \cdots, y_m\}$. Write $Z_k = {z_1,
 z_2, \cdots, z_k}$.  Then precision@k is defined as
 $|Z_k \cap Y|/k$, the fraction of true labels among the top k model
 predictions, and MAP@k is defined as the average of
 the precision@k' for each position $k'\leq k$ where
 $z_{k'} \in Y$.

While the softmax loss is a good
 surrogate to the rank loss~\cite{Zhang2004}, they are not
 identical. Indeed, the main contribution of this paper is to design
 an efficient adaptive sampling method for softmax model which achieves
 low rank loss (but not necessarily softmax loss).

\section{Algorithm}\label{sec:algo}

Our sampling algorithm works with the mini-batch stochastic gradient
descent method and is carried out for each batch. Consider a batch
$B=\{(x_i, y_i)\}$ of training examples. Algorithm~\ref{alg:tapas}
describes the two pass sampling algorithm. In the first pass, we
sample a subset $S'$ using the sampling distribution $Q$
just like in the sampled softmax, and in the
second pass we resample $S\subset S'$ adaptively dependent on the
batch $B$ and the current model parameters. We then use $S$ for
computing the gradient as done in sampled softmax.

The algorithm takes three parameters, where $n$ is the number of
output samples, $r\geq 1$ is the pre-sample factor, and $\tau$ is the
sampling temperature. Note that when $r=1$, TAPAS is identical to
sampled softmax.

\begin{algorithm}[tb]
   \caption{Two-pass adaptive sampling algorithm.}
   \label{alg:tapas}
\begin{algorithmic}
   \STATE {\bfseries Parameters:} $n$: number of samples, $r$: presmaple factor, $\tau$: sampling temperature.
   \STATE {\bfseries Input:} A batch $B$.
   \STATE {\bfseries Output:} $S\subset V$ where $|S|=n$.
   \STATE {\bfseries Non-adaptive pass}
   \begin{ALC@g}
   \STATE Sample $\min(r\cdot n, |V|)$ classes $S'$ according to sampling distribution $Q$.
   \end{ALC@g}
   \STATE {\bfseries Adaptive pass}
   \begin{ALC@g}   
   \STATE Compute and return \\ $\;\;\;S=\argmax^n_{y\in S'} \sum_{i\in B} \exp(\phi(x_i)\cdot \psi(y))/\tau)$.
   \end{ALC@g}
\end{algorithmic}
\end{algorithm}

\subsection{Non-adaptive sampling pass}

The first pass is non-adaptive sampling and can use any existing
sampling method $Q$. In our implementation, we use the standard squashed
empirical distribution as described in Section~\ref{sec:prelim}.  The
main purpose of the non-adaptive sampling pass is to obtain a sample
$S'$ with smaller size so the adaptive pass can be done
efficiently. The sampling also helps to reduce over-fitting as
explained later. In our experience, it seems a good tradeoff by
choosing the size of $S'$ in the order of $1\%$ to $10\%$ of the total
number of classes.

\subsection{Adaptive sampling pass}

In the adaptive sampling pass, a smaller set of samples $S$ are chosen
 from $S'$ dependent on $\{x_i\}_{i\in B}$ and the current model
 parameters $\phi,\psi$. Intuitively we choose the samples that are
 ``close'' to $x_i$, i.e. those classes $y$ such that
 $\phi(x_i)\cdot \psi(y)$ is large.  For efficiency, the sampling is
 done at the batch level. Intuitively we include a label $y$ in $S'$
 if it is close to $x_i$ for some $i\in B$.  We use temperature $\tau$
 to control the adaptivity --- when $\tau$ is smaller, the sampling is
 more adaptive as there is an increasing chance for $y$ to be selected
 if it is close to any of $x_i$. During the training, we reduce the
 temperature over time. This is similar to the intuition of decreased
 temperature in training softmax model~\cite{cesa1998finite}.

The adaptive sampling pass has a few benefits. First, it reduces the
sample size further by a factor of $r$. This reduces the gradient
computation cost significantly, and with the adaptive sampling,
hopefully does not lose much accuracy on the gradient estimation. Secondly, by using only the
examples that are close to the context, we focus on the ``hard
negatives'' and can obtain lower rank loss, similar to the intuition
in Wsabie~\cite{WestonBU11}. Thirdly, since the gradient update is on
a smaller set of classes, there is less chance for the gradient to
become stale, which is useful with asynchronous training. Since the
adaptive sampling deterministically chooses the $n$ classes close to
the context, the first pass is important for introducing randomness
into the sampling. Otherwise it may cause over-fitting of the
model. For example, at the extreme, if we always choose the top $n$
labels from the entire $V$, it would cause the model not to generalize
well as it may ``push away'' the correct labels. The adaptive sampling, while reducing the rank loss, actually leads to higher cross-entropy loss. This is due to that
the adaptive sampling skews the sampling distribution and hence results in a
more biased gradient estimation. However, in our algorithm, we intentionally does not correct for this skewness since we would like to emphasize on the ``hard negatives'' to improve the rank loss.

The adaptive sampling does come with a price. It requires to compute
the dot product $\phi(x_i) \cdot \psi(y)$ for all the pairs $i\in B$
and $y\in S'$. This computation can be expensive if the size of $S'$
is large. In the following, we describe a distributed approximation to
Algorithm~\ref{alg:tapas} that utilizes GPUs and the parallelism
supported by the distributed machine learning systems. With our
implementation, we show that TAPAS incurs a very low overhead.

\section{Implementation}\label{sec:impl}

We implemented TAPAS on Tensorflow~\cite{AbadiBCCDDDGIIK16}, but the
same algorithm can be easily adapted to the other distributed machine
learning platforms such as~\cite{torch,theano}.  In Tensorflow, the
machines are organized as \emph{workers} and \emph{parameter servers}
where the parameter server hosts the parameters such as the embeddings (in our case $\psi(y)$ for $y\in[V]$)
and neural network parameters, and the worker performs the gradient
computation and parameter update by communicating with the parameter
server. In the typical setup, the parameter servers are hosted on
CPUs, and the workers on GPUs for large scale training.

In the adaptive sampling pass, we need to compute $\argmax^n_{y\in
S'} \sum_{i\in B} \exp(\phi(x_i)\cdot \psi(y))/ \tau)$, which in turn
requires to compute $\phi(x_i) \cdot \psi(y)$ for each $i\in B$ and
$y\in S'$. If we use the standard setup of ``sample at worker,'' then
we would need to fetch the parameters $\psi(y)$ for $y\in S'$ from the
parameter server to the worker and performs the sampling at the
worker. When $S'$ is large, such method would incur large network
communication and cause significant slowdown of the training. There is
then less benefit from adaptive sampling. Instead, in our
implementation, we take the ``sample at the parameter server'' approach
by hosting the parameter servers on the GPUs and sampling on the
parameter servers.  In addition, to facilitate efficient distributed sampling, we
only approximately sample the top $n$ elements. More specifically, suppose there are $m$
parameter servers, and parameter server $\textrm{PS}_j$ hosts the
embeddings of a random subset $V_j\subseteq V$. After the pre-sample
$S'$ is obtained, $\textrm{PS}_j$ will only look at $S_j'=S'\cap V_j$
and select the top $n/m$ negative samples $S_j$. The union of $S_j$
will be an approximate top $n$ choice from $S'$. Our implementation is
described in Algorithm~\ref{alg:tapas-impl}.

\begin{algorithm}[tb]
   \caption{TAPAS.}
   \label{alg:tapas-impl}
\begin{algorithmic}
\STATE {\bfseries Worker}
\begin{ALC@g}
  \STATE Compute $\phi(x_i)$ for each $i\in B$.
  \STATE Sample $S'$ non-adaptively.
  \FOR{$j=1,\cdots,m$}
  \STATE Compute $S'_j = S' \cap V_j$
  \STATE Send $\{\phi(x_i)\}_{i\in B}$ and $S'_j$ to parameter server $j$.
  \ENDFOR
\end{ALC@g}
\STATE {\bfseries $\textrm{PS}_j$}
\begin{ALC@g}
  \STATE Compute \\ $\;\;\;S_j=\argmax^{n/m}_{y\in S_j'} \sum_{i\in B} \exp(\phi(x_i)\cdot \psi(y))/\tau)$.
  \STATE Send $S_j$ and corresponding parameters to the worker.
\end{ALC@g}  
\STATE {\bfseries Worker}
\begin{ALC@g}
  \STATE Compute the gradient with negative samples \\ $\;\;\;S=\cup_{j=1}^m S_j$.
  \STATE Update the parameters using the gradients.
\end{ALC@g} 
\end{algorithmic}
\end{algorithm}

In the implementation, $\{\phi(x_i)\}_{i\in B}$ is broadcast to all
the parameter servers. Compared to sampling at the worker, this does
incur some cost.  On the other hand, only those samples in $S$ are
sent from the parameter servers to the worker. Since the size of $S'$
(in the order of $100,000$) is typically much larger than the size of
$B$ (in the order of $1000$), Algorithm~\ref{alg:tapas-impl} has
significantly lower communication cost compared to sampling at the
worker. In addition, the sampling are distributed over the parameter
servers and can be done efficiently. One additional optimization is
that we do not compute the exact top $n$ elements in $S'$. Instead, we
take the union of the top $n/m$ elements from each of $m$ parameter
servers. So what we obtain is the approximate top $n$
samples. However, when the labels are randomly partitioned, the
approximation is fairly good as $n$ (in the order of $10,000$) is much
larger than $m$ (in the order of $10$ to $100$).

With the above implementation, the overhead of TAPAS is very small. In
our experiments, we do not observe much performance difference of
TAPAS with $n$ samples for $r=1$ (same as the sampled softmax) and for
$r=10$. But the rank loss for $r=10$ is significantly lower than
$r=1$, hence giving us a significant quality gain with similar
training time.

\section{Experiments}\label{sec:exper}
\subsection{Synthetic data}
We perform experiments on synthetic multi-class classification tasks
to test TAPAS on various combinations of parameters. We begin by
learning a linear classifier to approximate a standard Gaussian mixture with $1000$
classes. We then experiment with a large-scale synthetic dataset with $10000$
classes where the data is generated by a non-linear model. The second
setting seeks to imitate real-world modeling tasks.

\subsubsection{Linear Classifiers}\label{sec:synthetic_linear}
\paragraph{Data.} Our first experiment train linear classifiers on a
 Gaussian mixture dataset. We generate $V=1000$ random centroids
 $\mu_j\in\R^d$ from a Gaussian prior $\normal(0,\frac{c^2}{d}I_d)$,
 where we set $d=50$ and $c=3$. To generate data point $i$, we
 uniformly choose a label $y_i\in[V]$ and generate the associated
 position $x_i$ from the Gaussian distribution
 $\normal(\mu_{y_i},I_d)$. The goal is to learn a classifier that predicts
 $y$ given $x$.

\paragraph{Model.} Under this generative model, the true posterior of
 $\prob[y|x]$ has an exact softmax form
\begin{equation*}
  \prob[y=j|x]
  = \frac{\exp(w_j \cdot x+b_j)}{\sum_{j=1}^{V} \exp(w_j \cdot x+b_j)},
\end{equation*}
where $w_j=\frac{\mu_j}{\sigma^2}$ and
$b_j=-\frac{\|\mu_j\|^2}{2\sigma^2}$. For this simple problem, we can
compute the sample mean
$\what{\mu}_j=\frac{1}{|\{i:y_i=j\}|}\sum_{i:y_i=j}x_i$ and use it to
compute $w_j, b_j$. With a moderate number of samples, we could achieve
accuracy close to the information theoretical bound. However, that
requires seeing the whole dataset. Here we
are interested in evaluating the performance of stochastic gradient
based methods. We note even in this
simple setting, the convergence properties of the sampled softmax are not
fully understood.

In all the experiments we generate the training and test sets of size
$10^5$ and $10^4$, respectively, as described above. The model is a linear
softmax model. We use the {\sc AdaGrad} optimizer~\cite{DuchiHS2011} with
minibatches of size $B=16$. We use precision at $1$ to measure the
quality of each experiment.  We carry out two sets of experiments, the first
varying the number of samples $n$, and the second varying the
pre-sample factor $r$.
\begin{enumerate}[(1)]
\item Fix $n=16$ and let $r=1,2,4,8$. 
\item Fix pre-sample size $n\times r=128$, and let $n=16, 32, 64, 128$.
\end{enumerate}

We also run the full softmax computation for comparison. 

\begin{figure*}[htb]
  \centering
  \begin{subfigure}[b]{0.4\textwidth}
    \centering
    \includegraphics[width=\linewidth]{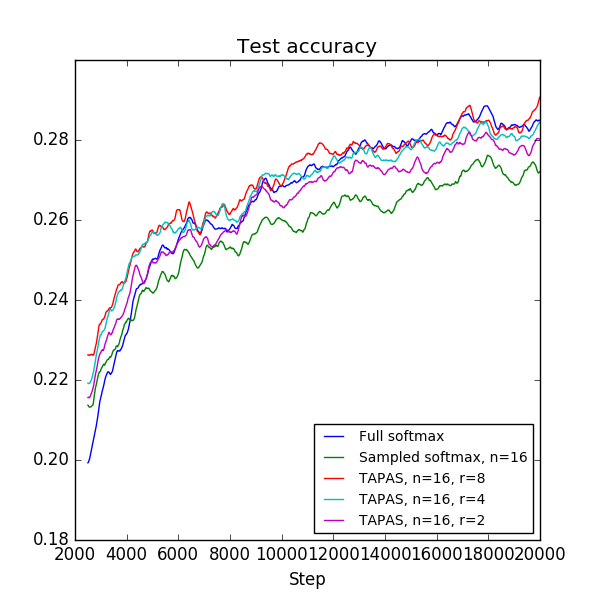}
    \caption{}
    \label{fig:linear_classifier_left}
  \end{subfigure} %
  \begin{subfigure}[b]{0.4\textwidth}
    \centering
    \includegraphics[width=\linewidth]{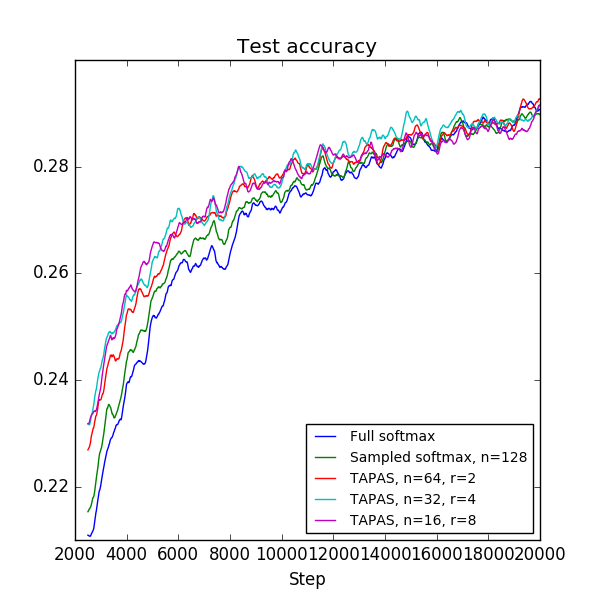}
    \caption{}
    \label{fig:linear_classifier_right}
  \end{subfigure}
  \caption{Results on linear classifiers. Left: fixing $n=16$, varying $r$. Right: fixing $nr=128$, varying $(n,r)$. We add a $\ell_2$-regularization of level $0.001$ in all cases to prevent over-fitting. We truncate the first $2500$ steps and smooth the curves by a moving average filter.}
  \label{fig:linear_classifier}
\end{figure*}

\paragraph{Results.} Figure~\ref{fig:linear_classifier} plots the test
 prediction accuracy of the classifiers of the two sets of
experiments.  Since the linear problem is relatively easy, we do not
observe big differences between different experiments.  For example,
there is less than $10\%$ difference between the full softmax and the
sampled softmax with $n=16$. However, even for such a simple problem,
there is visible quality difference with different sampling
strategies. In Figure~\ref{fig:linear_classifier}(a), we can see
for fixed $n$, increasing $r$ increases the accuracy, especially for
smaller $r$'s. From Figure~\ref{fig:linear_classifier}(b) we observe that the
accuracy of different experiments are getting close over time,
verifying that the accuracy has strong dependency on $n\cdot r$ so using
TAPAS with $n$ samples and pre-sample factor $r$ performs similarly to
the sampled softmax with $n\cdot r$ samples.  However, the latter is
much more expensive to run as it needs $r$ times more samples in the
gradient computation.

\begin{figure*}[htb]
  \centering
  \begin{subfigure}[b]{0.3\textwidth}
    \centering
    \includegraphics[height=1.6in]{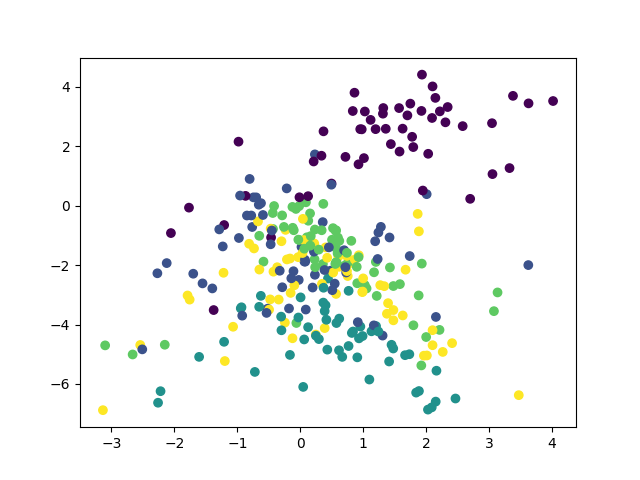}
    \caption{}
    \label{fig:nn_scatter}
  \end{subfigure}
  \begin{subfigure}[b]{0.3\textwidth}
    \centering
    \includegraphics[height=1.6in]{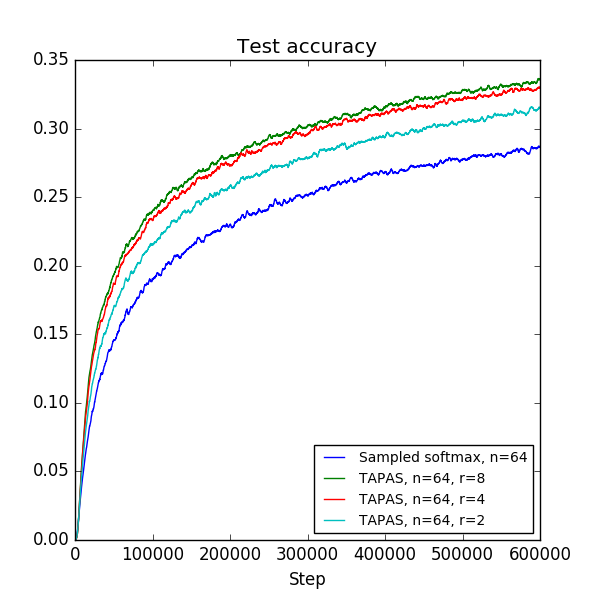}
    \caption{}
    \label{fig:nn_left}
  \end{subfigure} %
  \begin{subfigure}[b]{0.3\textwidth}
    \centering
    \includegraphics[height=1.6in]{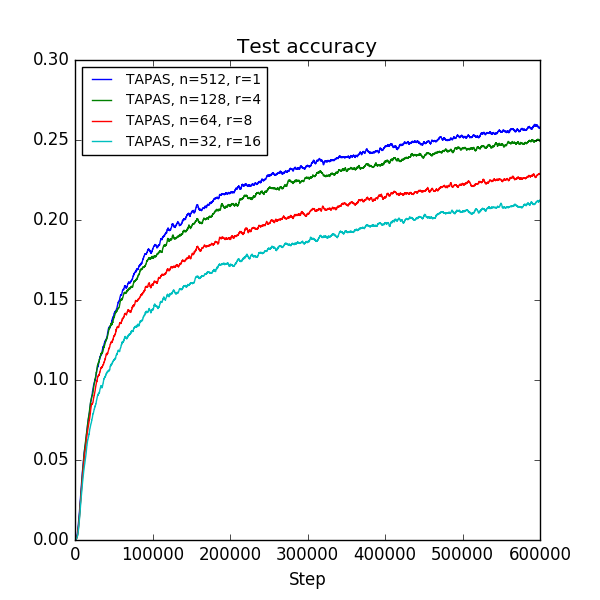}
    \caption{}
    \label{fig:nn_right}
  \end{subfigure}
  \caption{Results on two-layer networks. Left: an illustration of the
    generative model in two dimensions with 300 points, 5 classes, and
    generative hidden dimensions $(d_c,d_n,d_h)=(5,5,10)$. Middle and
    right: test accuracies of classifiers trained with TAPAS. Middle:
    fixing $n=64$, varying $r$. Right: fixing $nr=512$, varying
    $(n,r)$. We smooth the curves by a moving average filter.}
  \label{fig:nn}
\end{figure*}

\subsubsection{Neural networks}

We create a more challenging task by adding more classes and using a non-linear generative
model.

 \paragraph{Data.} We use a non-linear generative model to generate
 the data points as follows.  First, generate $V=10000$
 ``centroids'' $\mu_j\in\R^{d_c}$ similar to
 Section~\ref{sec:synthetic_linear}. To generate a data point $x_i$,
 we first choose the label $y_i\in[V]$ uniformly at random and form a
 input vector $\widetilde{x}_i=[\mu_{y_i}, z_i]\in\R^{d_c+d_n}$, where
 $z_i\sim\normal(0,\sigma^2I_{d_n})$ is a random Gaussian. This vector
 is then passed through a neural network to get $x_i$. The generative
 network has two layers where the first layer is a $d_h$-unit layer
 with ReLU activation and the second layer is a linear layer with $d$
 units.  Both layers are fully connected. This network imitates the
 generative neural networks~\cite{goodfellow2014generative}.  The
 weights of the generative networks
 are random Gaussians with proper scaling. We use parameters
 $d_c=d_n=10$, $d_h=50$, and $d=25$. For illustration, we plot two
 dimensional examples generated by such a method in
 Figure~\ref{fig:nn}(a). As can be seen, the generated
 clusters have significant overlapping and are more challenging to
 classify.

\paragraph{Model.} The classifier is a neural network with one
$50$-dimensional hidden layer, 
so both $x$ and $y$ gets embedded in $50$ dimensional space.

We create a train set of size $10^6$ and test set of size $10^5$. We
carry out the experiments with similar set up as in
Section~\ref{sec:synthetic_linear} with slightly larger parameters:
the batch size is set to $32$, and for the first set of experiments
$n=64$ and for the second set $n\cdot r=512$.

\paragraph{Results.} Figure \ref{fig:nn} shows the results of the two
set of experiments. For fixed $n$ as shown in
Figure~\ref{fig:nn}(b), we observe the similar phenomenon that the
accuracy increases when we increase $r$, except that the effect is
much more visible in this more challenging task. In
Figure~\ref{fig:nn}(c), with large sample size, there is a gain
of accuracy. However the accuracies of $n=512, r=1$ and $n=128, r=4$ are
almost the same, which still shows a significant gain as the latter
is 4 times faster in the gradient computation.

\subsection{Real data}

We apply TAPAS to a large scale classification problem with a real
data set. The data set consists of sequences of users' consumption of
items on a popular video site. Our training data consists of $200$
million sequences with average length of $30$, and the item comes from
a dictionary of size $500,000$. We build a softmax sequence model
for predicting the next five items in the sequence from the prefix of
the sequence. The scoring is done using the mean average precision
(MAP) at 20. The testing is done on $20$ million holdout
sequences. This is a fairly challenging task as the items have a
rather long tail distribution, for example, the top $20\%$ most popular items only occurs about $60\%$ of times. Our model is a complex neural network
model that achieved the highest precision result compared to
multiple internal implementations of the state-of-the-art methods.

In our experiments, we fix the model architecture and training
hyper-parameters but vary the pre-sample factor and the number of
samples. We experiment the combination of $n=1000, 8000$ and $r=1, 8$.
In all the experiments, the training is done using $6$ workers and $6$
parameter servers, all hosted on GPUs.

\paragraph{Efficiency.} Table~\ref{tbl:exp-steps} shows the number of
training steps per second for each experiment. We observe that for the
same $n$, increasing $r$ from $1$ to $8$ only causes a small, about $10\%$,
overhead. Increasing the value $n$ however does slow down the training
significantly.

\begin{table}[t]
\caption{Number of steps per second.}
\label{tbl:exp-steps}
\vskip 0.15in
\begin{center}
\begin{tabular}{l|c}
Experiment & Steps/Sec\\
\hline
$n=1000, r=1$ & $181$ \\
\hline
$n=1000, r=8$ & $173$ \\
\hline
$n=8000, r=1$ & $92$ \\
\hline
$n=8000, r=8$ & $81$ \\
\hline
\end{tabular}
\end{center}
\vskip -0.1in
\end{table}

\begin{figure}[h!]
  \centering
  \includegraphics[width=0.7\columnwidth]{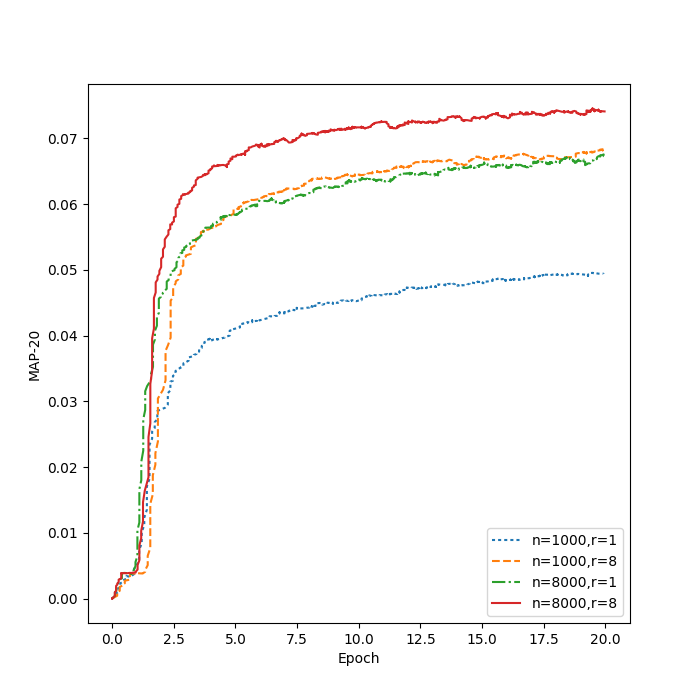}
  \caption{Results on MAP score. $n$: number of samples; $r$: pre-sample factor.}
  \label{fig:real-map20}
\end{figure}

\paragraph{MAP score.}  Figure~\ref{fig:real-map20} shows the
MAP@20 scores. As can be seen from the plot, with TAPAS, the MAP score
is greatly improved. When $n=1000$, the MAP score is improved from
$0.050$ to $0.068$ by increasing $r$ from $1$ to $8$, representing
almost $30\%$ improvement, and when $n=8000$, the improvement is
smaller, but still about $12\%$ from $0.067$ to
$0.075$. Interestingly, TAPAS with $n=1000, r=8$ has slightly higher
accuracy than $n=8000, r=1$.

\begin{figure}[h!]
  \centering
  \includegraphics[width=0.7\columnwidth]{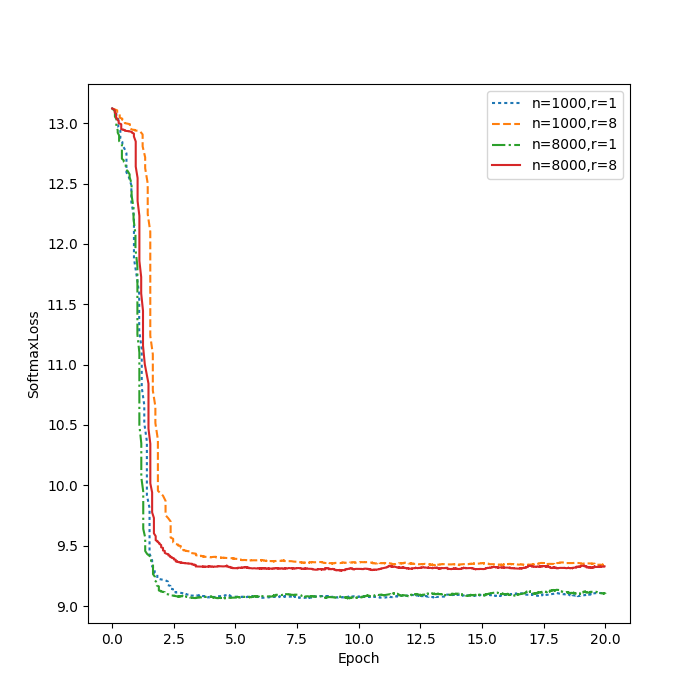}
  \caption{Results on softmax loss. $n$: number of samples; $r$: pre-sample factor.}
  \label{fig:real-softmax}
\end{figure}

\paragraph{Softmax loss.} As we described earlier, TAPAS suits
well on rank loss but it may not work so well on full softmax
loss. Indeed, Figure~\ref{fig:real-softmax} shows that while
the model trained with $n=8000,r=8$ has much higher MAP20 score than
the combination of $n=1000,r=1$ ($0.075$ vs $0.050$), it actually has
a slightly higher full softmax loss ($9.34$ vs $9.10$).

These experiments demonstrate that TAPAS can make large improvement on
the rank loss with fairly low overhead. On the other hand, it does not
improve on the full softmax loss.

\section{Conclusion}\label{sec:concl}
We present TAPAS as an effective sampling strategy for softmax
model. We have implemented the algorithm and plan to release the code
for open-source use. We are particularly thrilled by its effectiveness
on the rank loss. It remains an interesting question to establish a
formal connection between TAPAS sampling strategy and the rank loss,
probably under suitable assumptions.

\section*{Acknowledgments}
We would like to thank Zhifeng Chen for the idea of co-locating
sampling with the parameter sever and the help to make it work; Kunal
Talwar for useful discussion; Chris Colby and George Roumpos for the
open source tensorflow implementation; and Walid Krichene for many
useful comments on the paper.

\bibliography{tapas}
\bibliographystyle{icml2016}

\end{document}